\documentclass[11pt]{article}

\usepackage[final]{acl}

\usepackage{times}
\usepackage{latexsym}
\usepackage{amsmath}
\usepackage{amsfonts}

\usepackage[T1]{fontenc}

\usepackage[utf8]{inputenc}

\usepackage{microtype}

\usepackage{inconsolata}

\usepackage{graphicx}
\usepackage{booktabs}
\usepackage{url}
\usepackage[most]{tcolorbox}

\title{Explainable Disentangled Representation Learning for Generalizable Authorship Attribution in the Era of Generative AI}

\author{\textbf{Hieu Man$^*$, Van-Cuong Pham$^*$, Nghia Trung Ngo$^*$,} \\
\textbf{Franck Dernoncourt$^\dagger$, Thien Huu Nguyen$^*$} \\
  $^*$University of Oregon, OR, USA, $^\dagger$Adobe Research, CA, USA \\
  \texttt{\{hieum, cuongp, nghian, thienn\}@uoregon.edu, franck.dernoncourt@adobe.com}}

\begin{document}
\maketitle

\begin{abstract}
Learning robust representations of authorial style is crucial for authorship attribution and AI-generated text detection. However, existing methods often struggle with content-style entanglement, where models learn spurious correlations between authors' writing styles and topics, leading to poor generalization across domains. To address this challenge, we propose Explainable Authorship Variational Autoencoder (EAVAE), a novel framework that explicitly disentangles style from content through architectural separation-by-design. EAVAE first pretrains style encoders using supervised contrastive learning on diverse authorship data, then finetunes with a Variational Autoencoder (VEA) architecture using separate encoders for style and content representations. Disentanglement is enforced through a novel discriminator that not only distinguishes whether pairs of style/content representations belong to the same or different authors/content sources, but also generates natural language explanation for their decision, simultaneously mitigating confounding information and enhancing interpretability. Extensive experiments demonstrate the effectiveness of EAVAE. On authorship attribution, we achieve state-of-the-art performance on various datasets, including Amazon Reviews, PAN21, and HRS. For AI-generated text detection, EAVAE excels in few-shot learning over the M4 dataset. Code and data repositories are available online\footnote{\url{https://github.com/hieum98/avae}} \footnote{\url{https://huggingface.co/collections/Hieuman/document-level-authorship-datasets}}.
\end{abstract}

\section{Introduction}
Authorship Analysis, which identifies the stylistic fingerprints of authors, has become a critical technology for navigating modern digital landscape. A primary task within this field is Authorship Attribution (AA), which seeks to identify the author of a given text from a pool of candidates, with applications in intellectual property protection, academic integrity, and forensic investigation \cite{stover2016computational, stamatatos-2017-authorship}. Recent advancement of highly fluent Large Language Models (LLMs) has intensified these challenges while creating new tasks, such as AI-generated text detection, where the goal is to distinguish human-written content from machine-generated text \cite{10.1145/3531146.3533088,hazell2023spearphishinglargelanguage}.

\begin{figure}[t]
    \centering
    \includegraphics[width=0.45\textwidth]{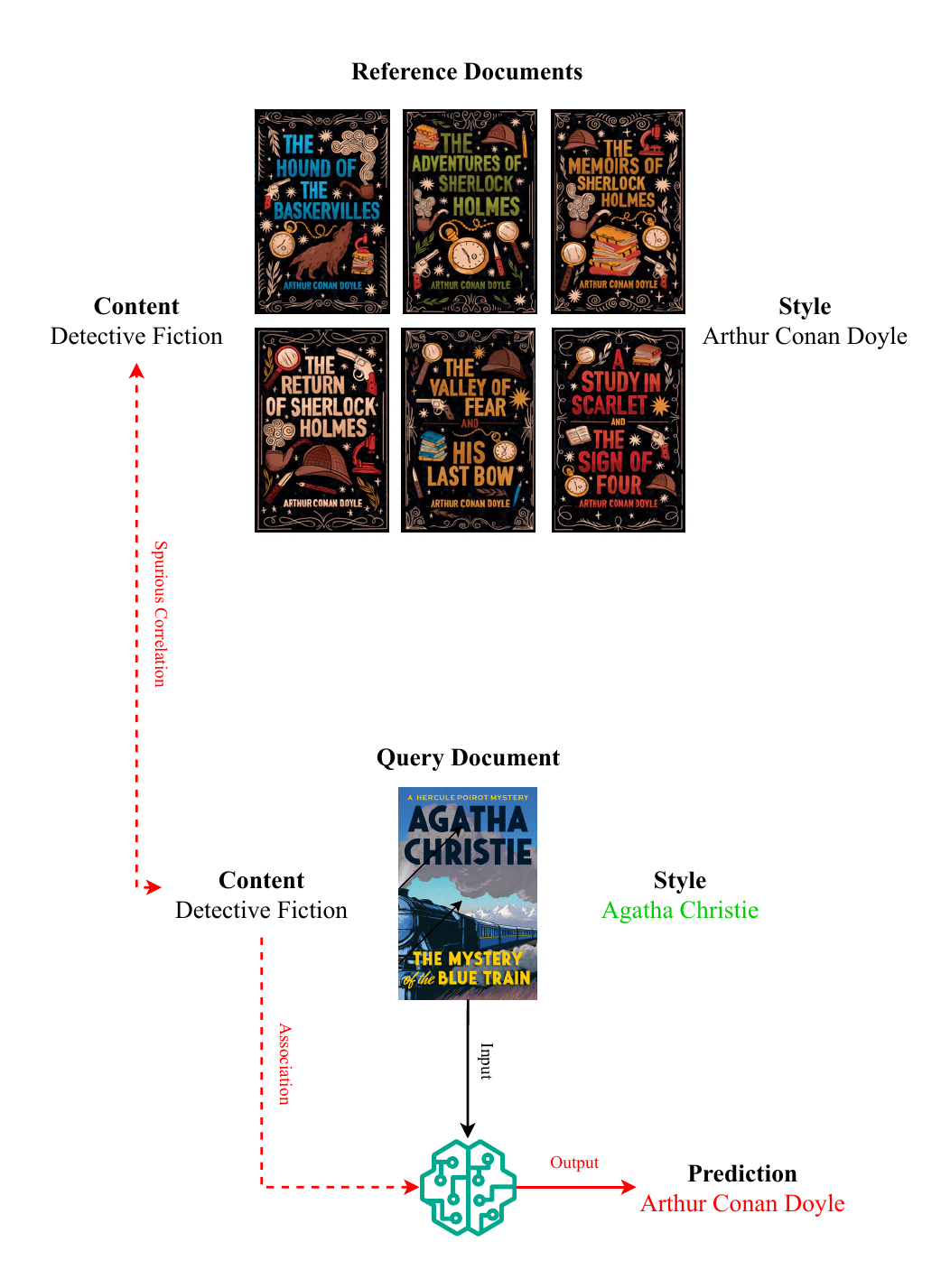}
    \caption{An example of content-style entanglement.}
    \label{fig:entanglement}
\end{figure}

The evolution of authorship attribution methods reflects broader trends in natural language processing. Early approaches relied on hand-crafted stylometric features and traditional machine learning classifiers \cite{10.5555/1527090.1527102, stolerman2014breaking, stamatatos-2017-authorship}. Though interpretable, they often struggled with scalability and domain transferability. Recent work has embraced neural approaches that learn representations directly from text using deep learning techniques, particularly contrastive learning \cite{boenninghoff2019explainable, rivera-soto-etal-2021-learning}. Despite significant progress in authorship analysis, a fundamental challenge known as the content confounding problem continues to limit the robustness and generalizability of these methods. This issue, formally studied as \emph{topic confusion}, occurs when models learn spurious correlations between authors' identities and the topics they frequently discuss, rather than capturing their intrinsic, topic-agnostic writing style. An intuitive example is presented in Figure \ref{fig:entanglement}, in which a model incorrectly labels a query document as a work by Sir Arthur Conan Doyle due to having learned to associate the author's identity with Detective Fiction content instead of author's unique writing style.

Current style-content disentanglement methods \cite{altakrori2021topic, sawatphol-etal-2022-topic, 10.1145/3626772.3657956} often rely on training a single encoder with contrastive learning objectives to learn holistic text representations and may implicitly conflate style and content into a single embedding. Moreover, they tend to employ small language model (SLM) encoders that have limited capacity to learn complex representations. As a result, these approaches struggle to generalize across diverse topics and domains. Moreover, none of the existing methods provides interpretability for the learned representations, making it difficult to understand what stylistic features are being captured and how they contribute to attribution decisions.

To address these limitations, we propose Explainable Authorship Variational Autoencoder (EAVAE), a novel framework that explicitly disentangles authorial style from content through architectural separation-by-design. EAVAE employs a two-stage training approach: First, we pretrain a style encoder based on LLMs using supervised contrastive learning on diverse authorship data to establish strong foundational representations; Second, we propose a novel finetuning framework with separate encoders for style and content representations, assuming style-content independence. Our framework utilizes Variational Autoencoder (VAE) to reconstruct the documents from their encodings. Moreover, we introduce an explainable discriminatory objective that not only distinguishes whether pairs of style/content representations originate from the same authors/content sources, but also generates natural language explanation for their decision. This dual objective of disentanglement and explainability explicitly alleviates confounding information and enhances model interpretability, providing insights into the features that contribute to the learned representations. Extensive experiments show that EAVAE achieves substantial improvements on the benchmark datasets Amazon Reviews \cite{ni-etal-2019-justifying}, PAN21 \cite{10.1007/978-3-030-58219-7_25} and HRS corpus\footnote{\url{https://www.iarpa.gov/research-programs/hiatus}} for authorship attribution, and strong performance on the M4 dataset \cite{wang2024m4multigeneratormultidomainmultilingual} for AI-generated text detection, demonstrating the generalization and robustness of authorial style representations from our architectural disentanglement mdethod.

Our contributions are threefold: (1) We introduce a disentangled representation learning framework specifically designed for robust authorship attribution across content domains with VAE; (2) We propose an explainable adversarial discriminatory objective that enforces disentanglement while also providing interpretable explanation for learned representations; and (3) We demonstrate state-of-the-art performance on challenging AA benchmarks and competitive results on {\it zero-shot} AI-generated text detection.

\section{Methodology}
We present Explainable Authorship Variational Autoencoder (EAVAE), a novel framework that addresses the content-confounding problem through architectural disentanglement and adversarial training with explainable feedback. EAVAE employs a two-stage approach: (1) contrastive pre-training to establish strong authorial representations, and (2) VAE-based finetuning with separate encoders for style and content, enforced through an explainable discriminator to ensure effective disentanglement and interpretability.

\subsection{Problem Formulation}
\label{sec:problem_formulation}
Following \cite{rivera-soto-etal-2021-learning, 10.1145/3626772.3657956}, we formulate authorship attribution as a ranked retrieval problem. Given a query document $d_q$, we aim to learn a function $f_\theta(d_q)$ that maps documents to representation vectors such that documents by the same author have higher cosine similarity than those by different authors:
\begin{equation}
\text{sim}(f_\theta(d_q), f_\theta(d_i)) > \text{sim}(f_\theta(d_q), f_\theta(d_j))
\end{equation}
where $\text{author}(d_q) = \text{author}(d_i) \neq \text{author}(d_j)$, and $\text{sim}(\cdot, \cdot)$ denotes cosine similarity. The key challenge lies in ensuring that $f_\theta$ captures authorial style patterns independently of content, avoiding spurious topic-author correlation that lead to poor cross-domain generalization.

\begin{figure}[t]
    \centering
    \includegraphics[width=0.45\textwidth]{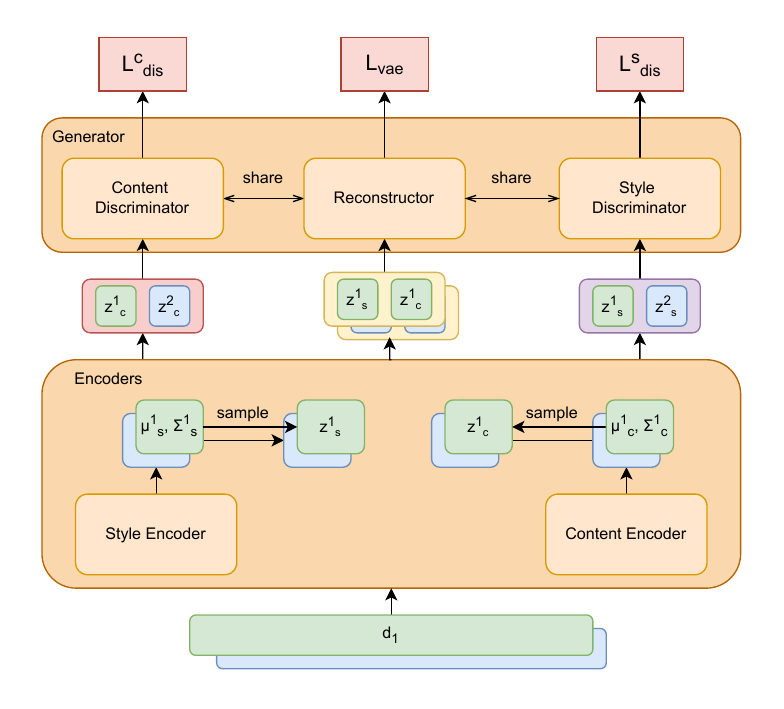}
    \caption{The architecture of Explainable Authorship Variational Autoencoder (EAVAE). EAVAE employs separate encoders for style and content, with an explainable discriminator that distinguishes whether pairs of style/content representations originate from the same or different authors/content sources, while generating natural language explanations for its decisions.}
    \label{fig:architecture}
\end{figure}

\subsection{Contrastive Pretraining}
\label{sec:pretraining}
We first establish strong authorial representations based on LLMs via supervised contrastive learning \cite{khosla2021supervisedcontrastivelearning} on a large, diverse authorship-labeled dataset. This pretraining stage aims to learn representations that cluster documents by the same author while separating those by different authors, providing a solid foundation for subsequent disentanglement. Formally, given a dataset $\mathcal{D} = \{(d_i, a_i)\}_{i=1}^N$ where $d_i$ is a document and $a_i$ is its author, we optimize:
\begin{equation}
\begin{aligned}
\mathcal{L}_{\text{con}} = -\sum_{i=1}^N \sum_{j \in \mathcal{P}(i)} \log \frac{\exp(r_i \cdot r_j / \tau)}{\sum_{k=1}^N \exp(r_i \cdot r_k / \tau)}
\end{aligned}
\end{equation}
where $r_i = f_\theta(d_i) / \|f_\theta(d_i)\|$ is the $\ell_2$-normalized representation, $\mathcal{P}(i) = \{k : a_k = a_i, k \neq i\}$ contains positive samples (same author), and $\tau$ is the temperature hyperparameter.

To enhance the contrastive learning process, following \cite{10.1145/3626772.3657956}, we employ hard negative mining using BM25 to mine the negative samples. Specifically, for each anchor document $d_i$, we retrieve the top-$K$ BM25 matches from different authors. This strategy forces the model to distinguish between documents that are lexically similar but stylistically different, reducing reliance on surface-level features. Additionally, we incorporate bidirectional attention mechanisms within the LLM, following recent advances in LLM's representation learning \cite{behnamghader2024llm2veclargelanguagemodels, muennighoff2025generativerepresentationalinstructiontuning}. This bidirectional context modeling significantly enhances the quality of learned representations by enabling the model to capture both forward and backward dependencies.

\subsection{Explainable Variational Autoencoder Fine-tuning}
\label{sec:finetuning}
The second stage explicitly disentangles style and content through a novel VAE architecture that combines architectural separation with adversarial training. As shown in Figure \ref{fig:architecture}, EAVAE employs dual encoders that map each document to separate style and content latent spaces, while explainable discriminators enforce disentanglement by learning to distinguish whether representation pairs originate from the same source. This design simultaneously achieves robust disentanglement and provides interpretable insights into the learned representations.

\subsubsection{Disentangled VAE Architecture}
EAVAE employs separate encoders for style and content to ensure that the learned representations effectively separate authorial style from content. The architecture consists of two encoders: $E_s(d) = (\mu_s, \sigma_s)$ for style and $E_c(d) = (\mu_c, \sigma_c)$ for content, where $\mu$ and $\sigma$ are the mean and standard deviation of the latent representations. The latent representations for style and content are sampled from multivariate Gaussian distributions $z_s \sim \mathcal{N}(\mu_s, \sigma_s^2)$ and $z_c \sim \mathcal{N}(\mu_c, \sigma_c^2)$, respectively. By separating the encoders, we encourage the model to learn distinct representations for authorial style and content. Formally, we make the style and content representations independent assumption:
\begin{equation}
q(z_s, z_c | d; E_s, E_c) = q(z_s | d; E_s) q(z_c | d; E_c)
\end{equation}
where $q(z_s | d; E_s)$ and $q(z_c | d; E_c)$ are the learned distributions for style and content, respectively. 
A shared reconstructor $G_{rec}(z_s, z_c)$ is then used to reconstruct the original document from the style and content representations. The VAE objective combines reconstruction fidelity with regularization:
\begin{equation}
\begin{aligned}
\mathcal{L}_{\text{vae}} = -\mathbb{E}_{z_s \sim q(z_s|d), z_c \sim q(z_c|d)}[\log p(d|z_s, z_c; G_{rec})] 
\\+ \beta_s \text{KL}(q(z_s|d) \| p(z_s)) + \beta_c \text{KL}(q(z_c|d) \| p(z_c))
\end{aligned}
\end{equation}
where $\beta_s$ and $\beta_c$ are hyperparameters that control the trade-off between the reconstruction loss and the KL divergence for style and content, respectively. The prior distributions $p(z_s)$ and $p(z_c)$ are typically chosen as standard normal distributions $\mathcal{N}(0, I)$. By employing separate encoders and enforcing independence between the learned distributions, EAVAE explicitly and effectively disentangles authorial style from content representations.

\subsubsection{Explainable Discriminator}
To further enforce the disentanglement of style and content representations, EAVAE employs an explainable discriminator that performs complementary tasks.
\textbf{Style Discrimination}: Given pairs of style representations $(z_s^i, z_s^j)$, it classifies whether they originate from the same author while generating explanations about distinguishing stylistic features.
\textbf{Content Discrimination}: Given pairs of content representations $(z_c^i, z_c^j)$, it classifies whether their topical content is similar while explaining the distinguishing content features.

Formally, given a pair of style/content representations sampled from latent distributions, the discriminator $G_{\text{expl}}$ is trained with the objective to encourage accurate classification with coherent explanations: 
\begin{equation}
\begin{aligned}
\mathcal{L}_{\text{dis}} = - \log p(o_s | z_s^i, z_s^j; G_{\text{expl}}) \\
- \log p(o_c | z_c^i, z_c^j; G_{\text{expl}})
\end{aligned}
\end{equation}
where $o_s$ and $o_c$ are ground-truth binary discrimination labels concatenated with target explanations for the style and content discrimination task, respectively. By employing a generator-based discriminator, EAVAE not only enforces the disentanglement of style and content representations but also provides natural language explanations for its decisions. This mechanism enhances the interpretability of the learned representations, allowing us to understand which stylistic and content features contribute to the model's decisions.

\begin{figure}[t]
    \centering
    \includegraphics[width=0.45\textwidth]{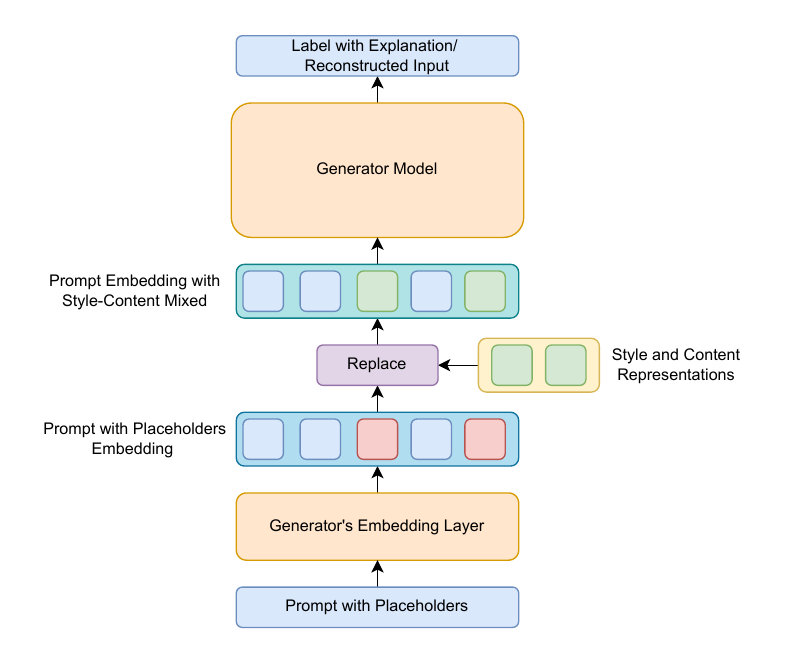}
    \caption{The architecture of unified generator with hybrid prompting mechanism.}
    \label{fig:generator}
\end{figure}

\subsubsection{Generator with Hybrid Prompting}
As shown in Figure \ref{fig:generator}, EAVAE employs a unified generative architecture with a hybrid prompting mechanism that serves dual purposes: document reconstruction from disentangled representations and discrimination with natural language explanations. This design eliminates the need for separate networks while enabling effective knowledge sharing across tasks through a sophisticated hybrid prompting mechanism. Formally, we formulate the task of reconstruction and discrimination as a conditional text generation problem. Given a pair of representations, i.e., $(z_s, z_c)$ for reconstruction or $(z_s^i, z_s^j)$ and $(z_c^i, z_c^j)$ for discrimination, the generator $G$ is trained to generate the target text $y$ (either the original document or the concatenated label and explanation) conditioned on the input representations and a task-specific prompt $p_t$:
\begin{equation}
\begin{aligned}
p(y | z, p_t; G) = \prod_{k=1}^{|y|} p(y_k | y_{<k}, z, p_t; G)
\end{aligned}
\end{equation}
where $y_k$ is the $k$-th token of the target text, and $y_{<k}$ are the preceding tokens. The task-specific prompt $p_t$ is designed with placeholders that are dynamically filled with representations, allowing the generator to adapt seamlessly to both reconstruction and discrimination tasks within a unified architecture. To do so, we employ a hybrid prompting mechanism that combines fixed template prompts with learnable soft prompts. Particularly, we first feed the prompt with placeholders $p_t$ into the generator's embedding layer to obtain the prompt embeddings $e_t=\{e_1, e_2, ..., e_{|p_t|}\}$ which includes the placeholder's token embeddings at positions $\{i, j\}$. We then replace these placeholder embeddings with the corresponding input representations, i.e., $e_i = z_s$ and $e_j = z_c$ for reconstruction, or $e_i = z_s^i$, $e_j = z_s^j$ and $e_i = z_c^i$, $e_j = z_c^j$ for discrimination. The modified prompt embeddings $\hat{e}_t$ are then fed into the generator to condition the text generation process.
\begin{equation}
\begin{aligned}
p(y | z, p_t; G) = \prod_{k=1}^{|y|} p(y_k | y_{<k}, \hat{e}_t; G)
\end{aligned}
\end{equation}
This hybrid prompting mechanism allows the generator to effectively leverage both task-specific guidance from the fixed template and flexibility from the textual representations, enabling it to perform both reconstruction and discrimination tasks seamlessly within a unified architecture. Besides, by replacing the placeholder token embeddings with the corresponding input representations, we ensure that position information is preserved, which is crucial for maintaining the contextual integrity in the embedded prompt.

\subsubsection{Training Objective}
The final EAVAE objective combines the reconstruction loss and the discrimination loss to ensure that the model learns to effectively separate authorial style from content while providing explainable insights into the learned representations. Formally, the overall loss function is defined as:
\begin{equation}
\begin{aligned}
\mathcal{L}_{\text{EAVAE}} = \mathcal{L}_{\text{vae}} + \lambda_{\text{dis}} \mathcal{L}_{\text{dis}}
\end{aligned}
\end{equation}
where $\lambda_{\text{dis}}$ is a hyperparameter that controls the trade-off between the reconstruction loss and the discrimination loss.

\section{Experiments}
This section presents our experimental setup. We first describe the training datasets in Section \ref{sec:train-dataset}, which consist of pretraining and fine-tuning datasets construction, followed by the evaluation tasks and benchmarks, including authorship attribution and AI-generated text detection in Section \ref{sec:evaluation} and the baselines in Section \ref{sec:baselines}. We then present our main findings in Section \ref{sec:results}, including the results on authorship attribution and AI-generated text detection, followed by ablation studies in Section \ref{sec:ablation_studies}. For details on implementation and choice of hyperparameters, please refer to Appendix \ref{sec:implementation_details}.

\subsection{Training Dataset}
\label{sec:train-dataset}
\textbf{Pretraining Dataset.} To facilitate supervised contrastive training, we aim to obtain a large, diverse, and representative corpus of text data with labeled authorship information for learning. To this end, we crawled data from various public sources and genres on the Internet, such as news articles, blogs, social media posts, and reviews. In addition, we introduced a series of preprocessing steps to clean and improve the quality of the collected data. Among others, we filtered out the documents that are less than 32 or longer than 512 tokens to preserve sufficient context for representation learning. We also retained only the authors who published between 10 and 1000 documents to ensure balanced contexts of authors for learning. Finally, we deduplicated the dataset to remove near-duplicate documents that could bias the learning process. The final pretraining dataset contains 27.4 million documents from 1.3 million unique authors across diverse topics and styles.

\textbf{EAVAE Fine-tuning Dataset.} For EAVAE fine-tuning, we employ a strategic hard pair mining approach to curate challenging training examples from the pre-training dataset. Following \cite{10.1145/3626772.3657956}, we identify two critical types of hard pairs: 1) documents by the same author discussing different topics with low content similarity, and 2) documents on similar topics by different authors with high content similarity. This hard pair mining strategy forces the explainable discriminator to learn robust disentangled representations that resist spurious style-content correlations. The mining process leverages GTE-Qwen2-1.5B \cite{li2023towards} to compute semantic document embeddings, followed by K-means clustering ($k=1000$) to establish topic structure. We then employ QwQ-32B \cite{qwq32b} to generate detailed natural language explanations for each mined pair and classification label, articulating specific stylistic and content-based evidence. After these steps, we obtain the final fine-tuning dataset containing 132k document pairs from 12k unique authors across diverse topics, along with binary labels for both style and content comparison tasks, and comprehensive explanations of these labels.

\subsection{Evaluation Tasks}
\label{sec:evaluation}
We evaluate EAVAE on two key tasks: authorship attribution and AI-generated text detection. 

\textbf{Authorship Attribution.} We adopt the standard retrieval-based evaluation protocol \cite{rivera-soto-etal-2021-learning, altakrori2021topic, sawatphol-etal-2022-topic, 10.1145/3626772.3657956}, where the model ranks a pool of candidate authors according to the cosine similarity between their learned representations and a query document. Performance is measured using \emph{Mean Reciprocal Rank (MRR)} and \emph{Recall@8 (R@8)}, which assess the accuracy with which the correct author is retrieved. We consider two granularity levels: \textbf{Document-level Attribution}, where each document is evaluated independently, and \textbf{Author-level Attribution}, where multiple documents by the same author are aggregated into a unified representation, offering a more robust characterization of writing style.  

Following prior work \cite{rivera-soto-etal-2021-learning, altakrori2021topic, sawatphol-etal-2022-topic, 10.1145/3626772.3657956}, we evaluate author-level attribution on \textsc{Amazon Reviews} \cite{ni-etal-2019-justifying} and \textsc{PAN21} \cite{10.1007/978-3-030-58219-7_25}, while document-level attribution is assessed on the \textsc{HRS} corpus \cite{iarpaHIATUS}, which spans five heterogeneous domains: BoardGameGeek reviews\footnote{\url{https://boardgamegeek.com}}, Global Voices articles\footnote{\url{https://globalvoices.org}}, Instructables tutorials\footnote{\url{https://www.instructables.com}}, Stack Exchange Literature posts\footnote{\url{https://literature.stackexchange.com}}, and Stack Exchange STEM posts\footnote{\url{https://academia.stackexchange.com/questions/tagged/stem}}. The \textsc{HRS} setting is particularly challenging due to its topical diversity and substantial author overlap across genres. 

\textbf{AI-generated Text Detection.} For machine-text detection, we follow the few-shot protocol of \cite{soto2024fewshot}, where cosine similarity between the style representation of the candidate document and a small set of reference documents yields a score indicating authorship or machine provenance. Evaluation is conducted on the \textsc{M4} benchmark \cite{wang2024m4multigeneratormultidomainmultilingual}, which contains outputs from multiple LLMs across diverse domains, including scientific writing (ArXiv, PeerRead), instructional content (WikiHow), and encyclopedic text (Wikipedia).  

We use standardized partial area under the ROC curve (\emph{pAUC@k}), restricted to false alarm rates below 1\%, as the primary metric, following \cite{soto2024fewshot}. Results are reported under two setups: \textbf{Single-target Detection}, where the system distinguishes outputs from a specific generator (e.g., ChatGPT) using only $k$ in-distribution samples of that model's writing, and \textbf{Multi-target Detection}, where $k$ examples are provided for multiple generators, and query documents are matched to each candidate via similarity scores. This setting more closely reflects practical machine-generated text detection scenarios, where multiple language models may be active simultaneously.  

\subsection{Baselines}
\label{sec:baselines}
We compare EAVAE against several strong recent baselines for authorship representation learning and disentanglement. These include Style-Embedding \cite{wegmann-etal-2022-author}, which uses a Siamese network with contrastive loss to learn author embeddings; LUAR \cite{rivera-soto-etal-2021-learning}, which employs a supervised contrastive loss over large-scale authorship-labeled data, and \citet{10.1145/3626772.3657956}, which enhances authorship representation learning with hard negative mining using counterfactual interventions.

\subsection{Main Results}
\label{sec:results}
\begin{table}[t]
\centering
\resizebox{\columnwidth}{!}{
\begin{tabular}{l|cc|cc|cc}
\toprule
Models & \multicolumn{2}{c|}{Amazon Reviews} & \multicolumn{2}{c|}{PAN21} & \multicolumn{2}{c}{Avg.} \\
\midrule
~ & MRR & R@8 & MRR & R@8 & MRR & R@8 \\
\midrule
Style Embedding \cite{wegmann-etal-2022-author} & 60.9 & 72.9 & 11.9 & 18.3 & 36.4 & 45.6 \\ \midrule
LUAR \cite{rivera-soto-etal-2021-learning} & 93.4 & 95.7 & 60.1 & \textbf{66.2} & 76.8 & 81 \\ \midrule
\citealt{10.1145/3626772.3657956} & 93 & 96.8 & 47.3 & 54.9 & 70.2 & 75.9 \\ \midrule
\textbf{Contrastive Pre-training (Our)} & 94 & 96.1 & 57.9 & 61.2 & 76 & 78.7 \\ \midrule
\textbf{EAVAE (Our)}  & \textbf{97} & \textbf{99} & \textbf{61} & \textbf{66.2} & \textbf{79} & \textbf{82.6} \\
\bottomrule
\end{tabular}
}
\caption{Results on Amazon Reviews and PAN21 for Author-level Authorship Attribution.}
\label{tab:Author-level}
\end{table}

\begin{table*}[t]
\centering
\resizebox{\textwidth}{!}{
\begin{tabular}{l|cc|cc|cc|cc|cc|cc}
\toprule
Models & \multicolumn{2}{c|}{HRS1.1} & \multicolumn{2}{c|}{HRS1.2} & \multicolumn{2}{c|}{HRS1.3} & \multicolumn{2}{c|}{HRS1.4} & \multicolumn{2}{c|}{HRS1.5} & \multicolumn{2}{c}{Avg.} \\
\midrule
~ & MRR & R@8 & MRR & R@8 & MRR & R@8 & MRR & R@8 & MRR & R@8 & MRR & R@8 \\
\midrule
Style Embedding \cite{wegmann-etal-2022-author} & 10.3 & 15.3 & 11.4 & 15.9 & 8.1 & 16.2 & 10.1 & 18.5 & 9.9 & 14.1 & 10.1 & 16 \\ \midrule
LUAR \cite{rivera-soto-etal-2021-learning} & 53.1 & 73.9 & 22.9 & 34.1 & 11.7 & 20.6 & 28.4 & 40.2 & 30.1 & 40.2 & 29.2 & 41.8 \\ \midrule
\citealt{10.1145/3626772.3657956} & 50 & 61.8 & 32.2 & 39.1 & 33.9 & 43.1 & 29.3 & 37.3 & 37.5 & 42.9 & 36.6 & 44.8 \\ \midrule
\textbf{Contrastive Pre-training (Our)} & 54.3 & 64.2 & 27.9 & 43.6 & 50.9 & 62.4 & \textbf{33.2} & 44.1 & 39.5 & 49.2 & 41.2 & 52.7 \\ \midrule
\textbf{EAVAE (Our)}  & \textbf{64.7} & \textbf{89.2} & \textbf{44.5} & \textbf{65.9} & \textbf{53.4} & \textbf{80.9} & 32.2 & \textbf{54.3} & \textbf{41.5} & \textbf{70.7} & \textbf{47.3} & \textbf{72.2} \\
\bottomrule
\end{tabular}
}
\caption{Results on HRS corpus for Document-level Authorship Attribution.}
\label{tab:HRS}
\end{table*}

\textbf{Authorship Attribution.} Table \ref{tab:Author-level} reports author-level results on Amazon Reviews and PAN21. EAVAE attains 97.0\% MRR and 99.0\% Recall@8 on Amazon Reviews (+3.6/+3.3 vs. LUAR \cite{rivera-soto-etal-2021-learning}) and 61.0\% MRR and 66.2\% Recall@8 on PAN21, matching or surpassing prior bests. Our two-stage training offers benefits beyond scaling LLM encoders: the VAE fine-tuning adds +3.1 MRR over contrastive pretraining alone. Table \ref{tab:HRS} shows document-level attribution on HRS, a five-domain cross-topic benchmark, where EAVAE averages 47.3\% MRR and 72.2\% Recall@8, improving over \citealt{10.1145/3626772.3657956} by +10.7 MRR and +27.4 Recall@8 (relative >40\%), highlighting the value of architectural disentanglement under topic–author confounds. Across datasets, EAVAE consistently outperforms recent methods, with especially large gains over Style-Embedding \cite{wegmann-etal-2022-author}, which is vulnerable to content confounds. While LLM-based contrastive pretraining is already strong, EAVAE’s VAE stage delivers complementary gains (+3.0 MRR on average author-level tasks; +6.1 MRR on document-level), validating that explicitly disentangling style from content improves robustness and cross-domain generalization in real-world authorship attribution.

\begin{table*}[t]
\centering
\resizebox{\textwidth}{!}{
\begin{tabular}{l|ccc|ccc|ccc|ccc|ccc}
\toprule
Models & \multicolumn{3}{c|}{ArXiv} & \multicolumn{3}{c|}{PeerRead} & \multicolumn{3}{c|}{WikiHow} & \multicolumn{3}{c|}{Wikipedia} & \multicolumn{3}{c}{Avg.} \\
\midrule
~ & pAUC@1 & pAUC@5 & pAUC@10 & pAUC@1 & pAUC@5 & pAUC@10 & pAUC@1 & pAUC@5 & pAUC@10 & pAUC@1 & pAUC@5 & pAUC@10 & pAUC@1 & pAUC@5 & pAUC@10 \\
\midrule
\multicolumn{16}{c}{\textbf{Single-target Detection}} \\ \midrule
LUAR \cite{rivera-soto-etal-2021-learning} & 61.5 & 89.4 & 98.4 & 61.1 & 91.6 & 97.8 & 57 & 86.1 & 96.4 & 52.1 & 70.5 & 86.7 & 63.2 & 87.5 & 95.9 \\ \midrule
\citealt{10.1145/3626772.3657956} & \textbf{64.5} & \textbf{96.1} & \textbf{99.7} & 63.4 & 90.4 & 98 & 56.8 & 92.5 & 99.2 & 54.8 & 86.1 & 90.8 & 64.4 & 93.1 & 97.5 \\ \midrule
\textbf{Contrastive Pre-training (Our)} & 55.6 & 88.7 & 98.9 & 65.4 & 85.9 & 92 & 65.1 & 93.5 & 98.9 & 55.3 & 82.5 & 93.8 & 65.2 & 90.1 & 96.7 \\ \midrule
\textbf{EAVAE (Our)}  & 62.5 & 85.4 & 95.1 & \textbf{66.1} & \textbf{93.1} & \textbf{99} & \textbf{68.4} & \textbf{97.6} & \textbf{99.9} & \textbf{56.4} & \textbf{91.3} & \textbf{98.6} & \textbf{65.7} & \textbf{93.5} & \textbf{98.5} \\ \midrule
\multicolumn{16}{c}{\textbf{Multi-target Detection}} \\ \midrule
LUAR \cite{rivera-soto-etal-2021-learning} & 53.5 & 80.4 & 96.1 & 57.8 & 85.1 & 95.7 & 53.3 & 80 & 94.2 & 50.9 & 64 & 83.7 & 60 & 81.9 & 93.9 \\ \midrule
\citealt{10.1145/3626772.3657956} & \textbf{57.6} & 81.5 & 96.7 & 61.5 & 79.9 & 90.7 & 53.6 & 86.6 & 97.1 & 52.3 & 83.3 & 86.4 & 61.8 & 86.3 & 94.2 \\ \midrule
\textbf{Contrastive Pre-training (Our)} & 51.4 & \textbf{83} & \textbf{97.6} & 61.8 & 82 & 89.7 & 55.6 & 90 & 98.7 & 52.5 & 80.2 & 85 & 61.6 & 87 & 94.2 \\ \midrule
\textbf{EAVAE (Our)}  & 53.5 & 74.4 & 90.1 & \textbf{64.8} & \textbf{86.3} & \textbf{96.8} & \textbf{64.4} & \textbf{92.6} & \textbf{99} & \textbf{52.8} & \textbf{83.9} & \textbf{87.4} & \textbf{62} & \textbf{87.4} & \textbf{94.7} \\
\bottomrule
\end{tabular}
}
\caption{Results on M4 for both Single-target and Multi-target AI-generated Text Detection.}
\label{tab:M4}
\end{table*}
\noindent \textbf{AI-generated Text Detection.} EAVAE demonstrates strong performance on AI-generated text detection across diverse domains and detection scenarios, even without any task-specific fine-tuning for the detection task. Table \ref{tab:M4} shows results on the challenging M4 benchmark, which spans multiple domains (ArXiv, PeerRead, WikiHow, Wikipedia) and evaluation settings. In single-target detection, where the system distinguishes outputs from a specific generator, EAVAE achieves an impressive average performance of 65.7\% pAUC@1, 93.5\% pAUC@5, and 98.5\% pAUC@10 across all domains, which represents improvements over \citealt{10.1145/3626772.3657956} and LUAR \cite{rivera-soto-etal-2021-learning} baselines. The multi-target detection setting, which better reflects real-world scenarios where multiple generators may be active simultaneously, reveals EAVAE's robustness in complex detection environments. EAVAE achieves a competitive average performance of 62.0\% pAUC@1, 87.4\% pAUC@5, and 97.7\% pAUC@10. The consistent improvement over baselines across most domains and metrics demonstrate that our architectural disentanglement approach can effectively capture the distinguishing patterns of AI-generated text, even without direct supervision for the detection task. This validates the generality and transferability of our learned style representations for the downstream task of AI-generated text detection.

\subsection{Ablation Studies}
\label{sec:ablation_studies}
To validate the contribution of each component in EAVAE, we conduct comprehensive ablation studies on the challenging HRS corpus for document-level authorship attribution. Table \ref{tab:ablation} presents the results, demonstrating that each architectural choice contributes to the overall performance.

Specifically, we consider the following ablations: \textbf{1. Contrastive Pre-training Only}, which removes the VAE fine-tuning stage and only utilizes the style encoder from contrastive pre-training; \textbf{2. W/o Disentanglement by Design}, which replaces the disentangled VAE architecture with a standard VAE that learns a single latent representation using a shared encoder (i.e., the style encoder from contrastive pre-training) and decoder; \textbf{3. W/o Explainable Discriminator}, which removes the explainable discriminator and only optimizes the VAE loss during fine-tuning; \textbf{4. MLP Discriminator}, which replaces the explainable discriminator with a simple MLP classifier that predicts authorship and topical labels without generating explanations; \textbf{5. Soft-Prompt Only}, which only uses learnable soft prompts to condition the generator without fixed template prompts. 

\begin{table}[t]
\centering
\resizebox{\columnwidth}{!}{
\begin{tabular}{l|cc}
\toprule
Models & \multicolumn{2}{c}{Avg.} \\
\midrule
~ & MRR & R@8 \\
\midrule
EAVAE (Full) & \textbf{47.3} & \textbf{72.2} \\ \midrule
Contrastive Pre-training Only & 41.2 & 52.7 \\ \midrule
W/o Disentanglement by Design & 44.5 & 58.3 \\ \midrule
W/o Explainable Discriminator & 45.4 & 66.0 \\ \midrule
MLP Discriminator & 45.5 & 65.4 \\ \midrule
Soft-Prompt Only & 43.3 & 66.1 \\ \bottomrule
\end{tabular}
}
\caption{Ablation studies on HRS corpus for Document-level Authorship Attribution. We report average MRR and Recall@8 across all five subsets.}
\label{tab:ablation}
\end{table}

First, the "Contrastive Pre-training Only" ablation, which omits the VAE fine-tuning stage, shows that while contrastive learning provides a strong foundation, the additional disentanglement and explainability mechanisms in EAVAE yield significant complementary benefits, as proven by both the ablation and main results. This confirms our hypothesis that combining large-scale contrastive pre-training with principled architectural disentanglement leads to more robust and generalizable authorship representations.

Architectural Disentanglement is the most critical component. The "W/o Disentanglement by Design" ablation, which uses a single encoder instead of separate style and content encoders, shows the largest performance degradation (MRR: 44.5\% vs. 47.3\%, R@8: 58.3\% vs. 72.2\%). This represents a loss of 2.8 MRR and 13.9 R@8 points, confirming that explicit architectural separation is crucial for learning robust authorial representations. The substantial R@8 improvement particularly highlights how disentanglement enables better ranking of candidate authors, which is critical for practical authorship attribution systems. 

Removing the explainable discriminator ("W/o Explainable Discriminator") results in performance drops of 1.9 MRR and 6.2 R@8 points, demonstrating its role in enforcing style-content independence. The comparison between different discriminator architectures reveals that our hybrid prompting approach significantly outperforms both traditional MLP discriminators (MRR: 45.5\%, R@8: 65.4\%) and soft-prompt only variants (MRR: 43.3\%, R@8: 66.1\%). Notably, the hybrid prompting mechanism provides particularly strong improvement (+4.0 MRR points over soft-prompt only), indicating its effectiveness in generating accurate explanations that guide the disentanglement process.

\section{Related Work}


Authorship attribution has moved from stylometry (function words, $n$-grams, shallow syntax with classical classifiers) to neural representation learning, where contrastive objectives deliver state-of-the-art results \cite{10.5555/1527090.1527102,stolerman2014breaking,stamatatos-2017-authorship,boenninghoff2019explainable,rivera-soto-etal-2021-learning,altakrori2021topic,sawatphol-etal-2022-topic, 10.1145/3626772.3657956}. Yet these systems are vulnerable to the \emph{content-confound problem}, formalized as \emph{topic confusion}, where models spuriously bind author identity to topic rather than style \cite{altakrori2021topic,sawatphol-etal-2022-topic,10.1145/3626772.3657956}.

One solution is \emph{implicit disentanglement} via supervised contrastive learning: LUAR and Contra-X cluster documents by author and curate content-matched pairs (e.g., within threads) so negatives are lexically similar but stylistically distinct \cite{rivera-soto-etal-2021-learning,ai-etal-2022-whodunit,wegmann-etal-2022-author,10.1145/3626772.3657956}. More recent work has further leveraged stylistic embeddings in an ensemble setting for cross-domain machine-generated text detection \cite{kandula-etal-2025-bbn}, demonstrating the value of combining complementary authorship signals. However, a single encoder typically conflates style and content, leaving residual topic leakage and limiting cross-domain transfer. A complementary solution is \emph{explicit disentanglement}, factorizing style and content using adversarial invariance and information-theoretic regularizers, as explored in style transfer and multilingual representation learning \cite{ganin2016domainadversarialtrainingneuralnetworks,park2021information,ramesh-kashyap-etal-2022-different,gao-etal-2023-learning-multilingual,wieting-etal-2023-beyond}. VAEs provide a principled route by imposing independence in latent variables, but standard text VAEs often require additional structure or objectives to realize clean factor separation \cite{kingma2022autoencodingvariationalbayes}. Our approach follows this explicit route with \emph{separation-by-design}: distinct encoders for style and content and an explainable discriminator that enforces independence while producing natural-language rationales.

A complementary line of work improves LLM-based text representations by enabling bidirectional context modeling in decoder-only LLMs \cite{behnamghader2024llm2veclargelanguagemodels, man-etal-2024-ullme, muennighoff2025generativerepresentationalinstructiontuning}. EAVAE inherits this paradigm for style encoding while going further by explicitly disentangling style from content through architectural separation and adversarial explainable training.

\section{Conclusion}
We present EAVAE, a novel framework for learning explainable disentangled representations of authorial style and content in text. By combining large-scale supervised contrastive pre-training with a disentangled VAE architecture and an explainable discriminator, EAVAE effectively separates stylistic and topical information while providing natural language explanations for its decisions. Extensive experiments on authorship attribution and AI-generated text detection demonstrate that EAVAE significantly outperforms strong baselines across diverse domains and scenarios. Ablation studies confirm the importance of each component, particularly architectural disentanglement and the explainable discriminator. Our results validate the core hypothesis that principled separation of style and content enables more robust and generalizable authorship representations. Future work could explore extending EAVAE to multilingual settings, leveraging recent advances in cross-lingual LLM embeddings \cite{man-etal-2025-lusifer}, incorporating additional stylistic dimensions such as sentiment or formality, and applying the framework to other modalities like code or speech.

\section*{Limitations}
While EAVAE demonstrates strong performance in disentangling authorial style and content, several limitations warrant consideration. First, while the explainable discriminator provides natural language explanations, the quality and interpretability of these explanations depend on the underlying language model's capabilities and may not always align with human intuition. Thus, further research is needed to enhance the fidelity and usefulness of generated explanations. Second, the current framework focuses primarily on binary authorship attribution and may require adaptation for multi-author or collaborative writing scenarios. Finally, while EAVAE shows promise in AI-generated text detection, evolving language models may produce outputs that increasingly mimic human style, potentially challenging the robustness of style-based detection methods over time. Addressing these limitations presents opportunities for future research to enhance the versatility and effectiveness of disentangled representation learning in authorship analysis.

\section*{Acknowledgments}

This research was partially supported by NSF Grant \#2239570. This research is also supported in part by the Office of the Director of National Intelligence (ODNI), Intelligence Advanced Research Projects Activity (IARPA), via the HIATUS Program contract 2022-22072200003. The views and conclusions contained herein are those of the authors and should not be interpreted as necessarily representing the official policies, either expressed or implied, of ODNI, IARPA, or the U.S. Government.

\bibliography{custom}
\clearpage

\appendix
\section{Pre-training Dataset Details}
Tabel \ref{tab:pretrain_dataset} summarizes the statistics of the pretraining dataset.

\begin{table*}[t]
\centering
\begin{tabular}{l|c|c|c}
\toprule
\textbf{Dataset} & \textbf{Num.Documents} & \textbf{Num.Authors} & \textbf{Avg.Length} \\
\midrule
Hackernews (HNI) \cite{baumgartner2020pushshiftredditdataset} & 2,066,399 & 27,449 & 127 \\
Stackexchange (SXT) \cite{baumgartner2020pushshiftredditdataset} & 2,267,326 & 63,499 & 151 \\
Twitter \cite{baumgartner2020pushshiftredditdataset} & 28,130 & 530 & 56 \\
New York Time Comments (NYT) \footnote{\url{https://www.kaggle.com/datasets/aashita/nyt-comments}} & 620,850 & 15337 & 130 \\
Amazon Product Reviews \cite{ni-etal-2019-justifying} & 3,509,764 & 69,327 & 226 \\
Blog Authorship Corpus \cite{10.1007/978-3-030-58219-7_25}& 326,228 & 7,575 & 235 \\
Yelp Reviews \footnote{\url{https://business.yelp.com/data/resources/open-dataset/}} & 1,809,220 & 60,861 & 165 \\
Reddit-Dump \cite{baumgartner2020pushshiftredditdataset} & 4,179,346 & 134,406 & 147 \\
Reddit Million User Dataset \cite{baumgartner2020pushshiftredditdataset} & 6,591,126 & 144,810 & 131 \\
IMDb \cite{seroussi2014authorship} & 77,447 & 1,628 & 215 \\
goodreads \cite{wan2019fine} & 2,066,232 & 251,021 & 128 \\
bookcorpus \cite{Zhu_2015_ICCV} & 132,508 & 66,256 & 509 \\
realnews \cite{zellers2019grover} & 425,940 & 212,970 & 753 \\
food.com-recipes \cite{zhang2021meta} & 1,327,536 & 212,834 & 61 \\
sfu-socc \cite{kolhatkar2020sfu} & 418,957 & 31850 & 96 \\
wiki-articles \footnote{\url{https://dumps.wikimedia.org/}} & 1,332,872 & 32,333 & 475 \\
\midrule
\textbf{Total} & \textbf{27,445,334} & \textbf{1,348,420} & \textbf{265} \\
\bottomrule
\end{tabular}
\caption{Statistics of the pretraining dataset used for supervised contrastive learning. The average length is measured in tokens.}
\label{tab:pretrain_dataset}
\end{table*}

\section{Implementation Details}
\label{sec:implementation_details}
We implement EAVAE using the HuggingFace Transformers library with PyTorch. During the pre-training stage, we initialize the supervised contrastive learning framework using Qwen2-1.5B \cite{yang2024qwen2technicalreport} as the backbone model. The contrastive learning phase employs a batch size of 512 with AdamW optimizer \cite{loshchilov2019decoupledweightdecayregularization} at learning rate $2e^{-4}$, training for 2 epochs using LoRA \cite{hu2021loralowrankadaptationlarge} with rank $r=16$ to significantly reduce memory footprint. The contrastive temperature parameter $\tau$ is set to 0.02. For the fine-tuning stage, we instantiate separate encoder components: an off-the-shelf GTE-Qwen2-1.5B \cite{li2023towards} serves as the content encoder, while the pre-trained model from the contrastive learning stage becomes the style encoder. The unified generator architecture employs Qwen2-1.5B \cite{yang2024qwen2technicalreport}. The complete EAVAE framework undergoes joint training with batch size 64 over 2 epochs using the AdamW optimizer at learning rate $1e^{-4}$. LoRA adaptation with rank $r=16$ is applied consistently across all components to maintain parameter efficiency. The VAE regularization hyperparameters are carefully tuned: $\beta_s = \beta_c = 0.1$ for style and content KL divergence terms, respectively, balancing reconstruction quality with disentanglement strength, while the discriminator loss factor $\lambda_{\text{dis}} = 0.5$ ensures effective adversarial training between the discriminator and encoders.

\section{Explainable Analysis}
To qualitatively assess the effectiveness of EAVAE in providing explainable disentangled representations, we present case study where the explainable discriminator generates natural language explanations for its classification decisions: 

\begin{tcolorbox}[colback=lightgray, colframe=black!20, breakable]
\textbf{Text 1:} This was just so-so for us. It was very bland. I made it exactly as written, but next time I would make some changes. I would spice up the chicken before I cooked it with some mexican spice mix, and I'd also add some sliced jalapeno's and maybe some cumin to the beans. I would make this again, but only with those changes. Made for PAC Spring 09.

\textbf{Text 2:} These cupcakes are to die for! The flavor is exactly what it says, pink lemonade....only in a cupcake! I added 1tbsp of fresh squeezed lemon juice to the batter. The only problem I had was getting them out of the cupcake pan without breaking them apart since they are very delicate. I topped them simply with cool whip. I will definately, yes definately make these again!! Made for Comfort Cafe Summer 09.

\textbf{Authorship Label:} Same author

\textbf{Style Explanation:} The two text samples exhibit several consistent stylistic features suggesting they were written by the same author. Both use colloquial expressions like \textbf{"to die for"} (Text 2) and \textbf{"so-so"} (Text 1), indicating a informal tone. The structure of reviewing recipes with personal modifications is shared: Text 1 mentions adding spices to chicken and beans, while Text 2 includes lemon juice and cooling topping. Both conclude with phrases like \textbf{"make this again"}, reinforcing recurrence patterns. Additionally, both entries include formatted notes, \textbf{"Made for..."}, maintaining a similar documentation style. 
\end{tcolorbox}

As shown in the case study, the explainable discriminator effectively generates clear, text-linked explanations that cite specific stylistic evidence for authorship comparisons. It identifies style beyond shared topic, highlighting parallel structure and recurring phrases (e.g., "make this again"), while separating stylistic cues from sentiment or content overlap to support the same-author inference. This qualitative analysis demonstrates EAVAE's capability to provide interpretable insights into the learned representations, enhancing trust and transparency in authorship attribution tasks.

\section{Prompting Details}
\label{sec:prompting}
We provide the detailed prompt templates used for the reconstruction and discrimination tasks in EAVAE below:

\textbf{Reconstruction Prompt Template:}
\begin{tcolorbox}[colback=lightgray, colframe=black!20, breakable]
Given the style representation and content representation of a text, reconstruct the original text.

Style representation: <placeholder>

Content representation: <placeholder>
\end{tcolorbox}

\textbf{Style Discrimination Prompt Template:}
\begin{tcolorbox}[colback=lightgray, colframe=black!20, breakable]
Given two style representations, determine if they are written by the same author or not. Your analysis should focus on stylistic features.

Your analysis should be concise but thorough, highlighting the most significant stylistic markers that support the given attribution.

Provide your analysis in valid JSON format with exactly two fields:

1. 'determination': 'same author' or 'different author'

2. 'explaination': Your analysis explaining why the texts appear to be written by the same author or different authors.

The style representations for Text 1, Text 2, respectively, are:

Text 1's style representation: <placeholder>

Text 2's style representation: <placeholder>
\end{tcolorbox}

\textbf{Content Discrimination Prompt Template:}
\begin{tcolorbox}[colback=lightgray, colframe=black!20, breakable]
Given two content representations, determine if they express the same core content, regardless of stylistic differences. 

Provide your analysis in valid JSON format with exactly two fields:

1. 'explaination': A concise explanation justifying your determination, highlighting key similarities or differences in content

2. 'determination': Either 'same content' or 'different content'

The content representations for Text 1, Text 2, respectively, are:

Text 1's content representation: <placeholder>  

Text 2's content representation: <placeholder>
\end{tcolorbox}

\end{document}